\documentclass{article}

\usepackage{arxiv}

\usepackage[utf8]{inputenc} 
\usepackage[T1]{fontenc}    
\usepackage{hyperref}       
\usepackage{url}            
\usepackage{booktabs}       
\usepackage{amsfonts}       
\usepackage{nicefrac}       
\usepackage{microtype}      
\usepackage{lipsum}

\usepackage{graphicx}
\graphicspath{ {./images/} }
\usepackage{xcolor}
\usepackage{tabularx}
\usepackage{longtable}
\usepackage{subcaption}
\usepackage{diagbox}
\usepackage{rotating}
\usepackage{booktabs}
\usepackage{hyperref}

\title{Benchmarking Multi-Scene Fire and Smoke Detection}

\author{
 Xiaoyi Han \\
 Zhejiang University, China\\
  \texttt{hanxiaoyicoder@gmail.com}\\
   \And
 Nan Pu \\
 University of Trento, Italy\\
 \texttt{nan.pu@unitn.it}\\
  \And
 Zunlei Feng \\
 Zhejiang University, China\\
 \texttt{zunleifeng@zju.edu.cn}\\
  \And
 Yijun Bei \\
 Zhejiang University, China\\
 \texttt{beiyj@zju.edu.cn}\\
  \And
 Qifei Zhang \\
 Zhejiang University, China\\
 \texttt{cstzhangqf@zju.edu.cn}\\
  \And
 Lechao Cheng\thanks{Corresponding author}\\
 Hefei University of Technology, China\\
  \texttt{chenglc@hfut.edu.cn} \\
 \And
 Liang Xue \\
 Suzhou City University, China\\
 \texttt{xliang@szcu.edu.cn}\\
}

\begin{document}
\maketitle
\begin{abstract}
The current irregularities in existing public Fire and Smoke Detection (FSD) datasets have become a bottleneck in the advancement of FSD technology. Upon in-depth analysis, we identify the core issue as the lack of standardized dataset construction, uniform evaluation systems, and clear performance benchmarks. To address this issue and drive innovation in FSD technology, we systematically gather diverse resources from public sources to create a more comprehensive and refined FSD benchmark. Additionally, recognizing the inadequate coverage of existing dataset scenes, we strategically expand scenes, relabel, and standardize existing public FSD datasets to ensure accuracy and consistency. We aim to establish a standardized, realistic, unified, and efficient FSD research platform that mirrors real-life scenes closely. Through our efforts, we aim to provide robust support for the breakthrough and development of FSD technology. The project is available at \href{https://xiaoyihan6.github.io/FSD/}{https://xiaoyihan6.github.io/FSD/}.\footnote{Code is available: \href{https://github.com/XiaoyiHan6/MS-FSDB}{https://github.com/XiaoyiHan6/MS-FSDB}.}

\end{abstract}


\section{Introduction}
\label{sec:Introduction}

Fire, a formidable and highly dangerous disaster, can result in substantial loss of life, property, societal upheaval, and adverse ecological consequences in various scenes, such as urban conflagrations~\cite{1,2,3}, forest fires~\cite{4,5}, industrial infernos~\cite{6,7}, and transportation incidents~\cite{8,9}. Thus, the development of an open source and standardized public Fire and Smoke Detection (FSD) benchmark is imperative given the profound impact and unpredictable nature of fires.\\

After reviewing past FSD datasets, we identify three key deficiencies: First, the limitation in image resolution (typically $<~300~\times~300$ pixels)~\cite{10} hampers the ability of advanced models to learn crucial features effectively. Second, most previous FSD datasets consist mainly of images extracted frame by frame from videos, leading to a substantial repetition of similar fire scenes across hundreds of images. Finally, the lack of standardization in labeling in previous datasets hinders their seamless integration into models, along with inconsistent evaluation criteria, as illustrated in Table~\ref{tab1}. Examples from the datasets are shown in the section  “Related Work.” \\

To tackle these challenges, we systematically standardize the previous public FSD dataset and establish unified evaluation criteria to ensure data consistency and fairness in evaluation. Recognizing the limitations in scene coverage of existing datasets, we extensively collect diversified resources from various open sources and conduct fire-related scene expansion to enrich the dataset's diversity and practical utility. Furthermore, to assess the generalization performance of our dataset, we conduct cross-validation experiments on multiple public FSD datasets for comparative testing. The results of these experiments strongly confirm the excellent generalizability and high applicability of our benchmark. In summary, the main~\textbf{contributions} are summarized as follows: 1) We built a new extensive FSD benchmark with 2,731 scenes. Based on our information, the dataset is the first FSD benchmark that includes smoke and flame targets in various scenes. 2) We compare our benchmark with the top five most popular FSD datasets by aligning with our data format through image screening and labeling. 3) To assess the generalization performance of various FSD datasets fairly, we establish a standardized independent testing set.

\begin{table}
\vspace{-2em}
\renewcommand\arraystretch{0.5}
\caption{The table displays attributes like name, year, element, number of samples, label, and limitations of previous fire datasets.}  
\label{tab1}
\begin{tabular}{lllllll}  
\toprule  
\multicolumn{2}{l}{The name of Dataset} & Year & Element & Samples & Label & Limitations\\  
\midrule 
\multicolumn{2}{l}{VisiFire\textsuperscript{\cite{11}}} & 2007 & video & 40 & None & stale,unvaried\\
\multicolumn{2}{l}{FIRESENSE\textsuperscript{\cite{12}}} & 2014 & video & 49 & None & dubious,superfluous\\
\multicolumn{2}{l}{BoWFire\textsuperscript{\cite{13}}} & 2015 & image & 466 & Inconvenient & selective,undiversified\\   
\multicolumn{2}{l}{Furg-Fire-Dataset\textsuperscript{\cite{14}}} & 2017 & video & 23 & Inappropriate & biased,homogeneous\\    
\multicolumn{2}{l}{\cite{15}}& 2019 & image & 72,012 & None & redundancy\\
\multicolumn{2}{l}{FD-dataset\textsuperscript{\cite{16}}} & 2020 & image & 50,000 & None & incomplete\\
\multicolumn{2}{l}{Fire-Smoke-Dataset\textsuperscript{\cite{10}}} & 2021 & image & 3,000 & None & pixelated \\
\multicolumn{2}{l}{\cite{17}} & 2021 & image & 33,666 & None & only-smoke\\
\multicolumn{2}{l}{UAV-Images-Dataset\textsuperscript{\cite{18}}} & 2022 & image & 49,452 & None & duplication\\
\multicolumn{2}{l}{DFS-dataset\textsuperscript{\cite{19}}} & 2022 & image & 9,462 & appropriate & cumbersome\\
\multicolumn{2}{l}{Forest-Smoke-Dataset\textsuperscript{\cite{20}}} & 2023 & image & 6,500 & unavailable & N/A\\
\bottomrule   
\end{tabular}
\end{table} 

\section{Related Work}
\label{sec:related work}
As depicted in Table~\ref{tab1}, Toreyin and Cetin et al.~\cite{11} introduce the VisiFire dataset in 2007, encompassing various video types including 13 flame videos, 21 smoke videos, 4 forest smoke videos, and 2 others. Despite its prominence as a leading Fire and Smoke Detection (FSD) dataset, VisiFire is characterized by limitations: absence of labeling and scene diversity, and relatively scant data volume compared to contemporary datasets. In 2014, Dimitropoulos et al.~\cite{12} present the FIRESENSE dataset, comprising 11 flame videos, 16 neg-fire videos, 13 smoke videos, and 9 neg-smoke videos. However, individual video files in this dataset entail substantial data volume, potentially introducing redundancy and compromising scene uniformity. The subsequent introduction of the BoWFire dataset~\cite{13} in the following year aims to address labeling and segmentation needs, emphasizing high image clarity. Despite encompassing a broader range of scenes than its predecessors, its overall data volume remains relatively modest, totaling 466 images. Notably, the training set includes 240 images sized at $50 \times 50$, possibly limiting the information content available for model training. In 2017, the Furg-Fire-Dataset~\cite{14} is introduced, featuring 23 videos with each single-frame image labeled. However, the suitability of these labels for existing detection models may be debatable. Khan et al. introduce a new dataset in 2019~\cite{15}, primarily focused on smoke detection and comprising 72,012 images across four categories. However, redundancy issues arise from the inclusion of cropped images from videos. In 2020, the FD-dataset~\cite{16} is introduced, consisting of 50,000 images solely for fire-related tasks and lacking smoke-related images, primarily utilized for image classification. Subsequently, in 2021, the Fire-Smoke-Dataset is presented~\cite{10}, encompassing 3,000 images for classification purposes but lacking localization labels, with some images exhibiting pixelation due to small sizes. In the same year, He et al.~\cite{17} propose a dataset containing 33,666 images, addressing smoke-related tasks but lacking flame images. In 2022, Almeida et al.~\cite{18} introduce the Dataset2-UAV Images of Wildfires, categorized into various areas, yet suffering from image redundancy issues. Also in 2022, the DFS-dataset~\cite{19} is introduced, offering a substantial number of images categorized by fire proportion but facing challenges due to redundant data and multiple watermarks. In 2023, the Forest-Smoke-Dataset~\cite{20} is introduced, focusing solely on forest smoke scenes, although accessibility issues persist.\\

Many FSD datasets prioritize classification tasks, leading to label scarcity for localization, which poses challenges for firefighters. Incorporating labels early in the dataset creation process can facilitate timely fire localization, potentially mitigating further fire spread and reducing associated risks~\cite{19}. However, opting for segmentation tasks may entail significant manual costs.\\

To address these challenges, we present a novel multi-scene FSD benchmark encompassing both classification and regression labels, featuring a diverse array of indoor and outdoor scenes meticulously curated to eliminate redundancy. Our benchmark not only facilitates flame detection but also smoke detection tasks, thus offering a comprehensive resource for fire-related investigations. By refining and standardizing existing public FSD datasets and establishing uniform evaluation criteria, we aim to significantly advance FSD research, fostering future endeavors in this field.\\

\section{Multi-Scene Fire and Smoke Detection Benchmark}
\begin{figure}[t]
     \subfloat[]{
    \begin{minipage}{0.45\linewidth}
    \includegraphics[width=1\textwidth]{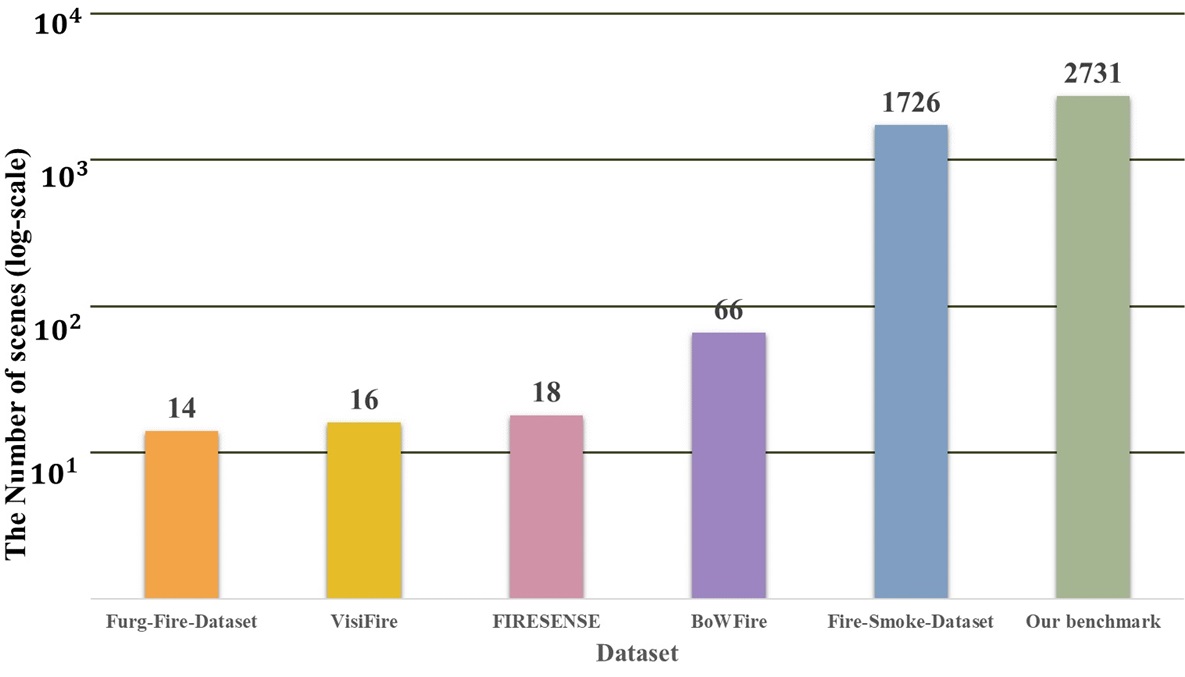}\vspace{1pt}
    \end{minipage}}
     \subfloat[]{
    \begin{minipage}{0.5\linewidth}
    \includegraphics[width=1\textwidth]{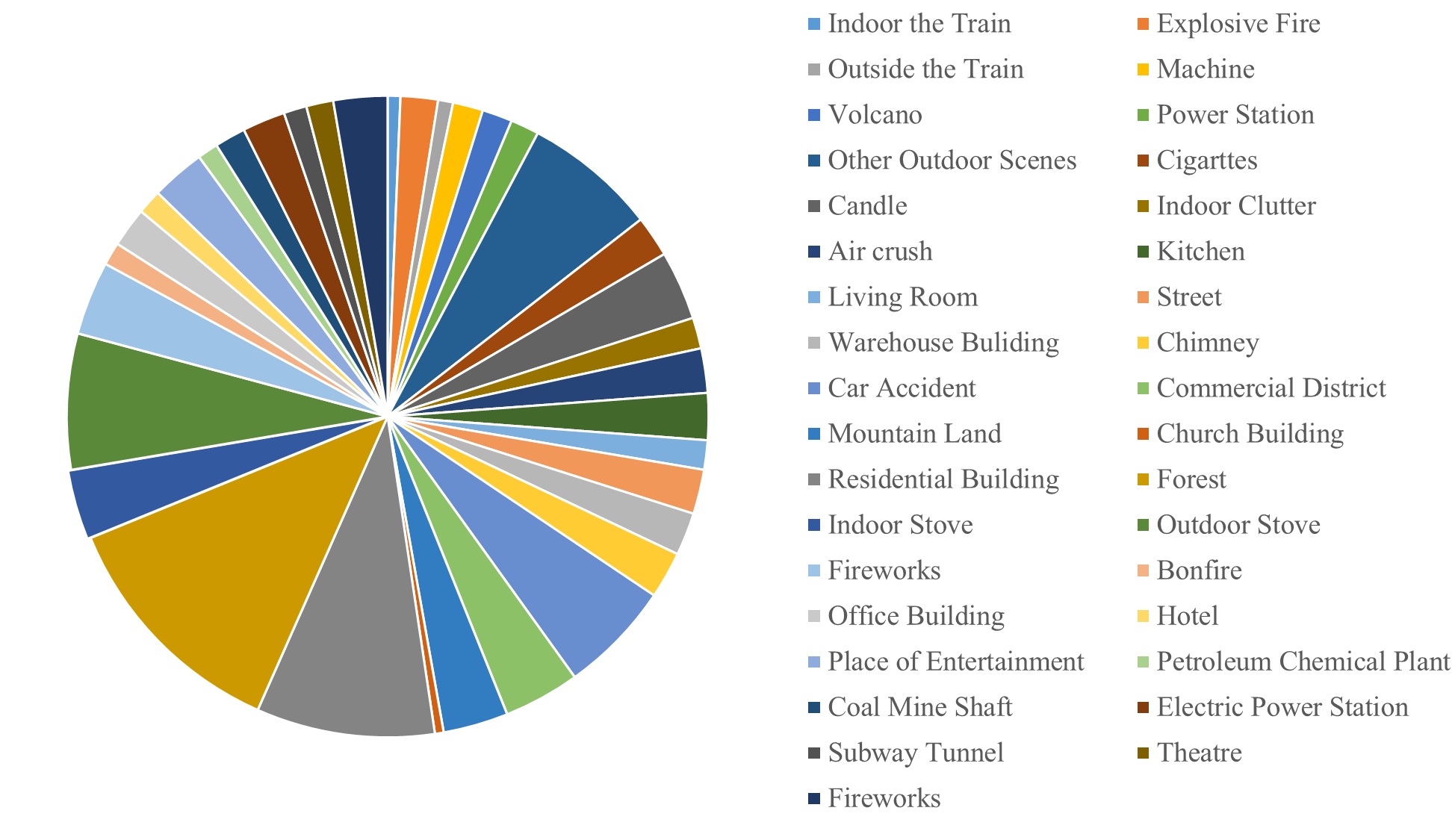}
    \end{minipage}}
\caption{Illustrations of several Fire and Smoke Detection (FSD) datasets statistics. (a) represents the statistics comparison of several FSD datasets scenes. (b) provides more detailed statistics for our benchmark. In (b), the area of the sector corresponds to the number of scenes. Additionally, the total number of scenes equals 2731, as displayed in (a).}
\label{fig1}
\vspace{-2em}
\end{figure}

\begin{figure}[t]
    \centering
    \subfloat[Electric Power Station]{
    \begin{minipage}{0.49\textwidth}
    \includegraphics[width=1\linewidth]{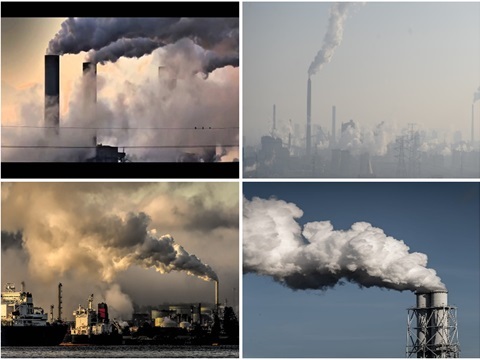}\vspace{0.5pt}
    \end{minipage}}
    \subfloat[Indoor Clutter]{
    \begin{minipage}{0.49\textwidth}
    \includegraphics[width=1\linewidth]{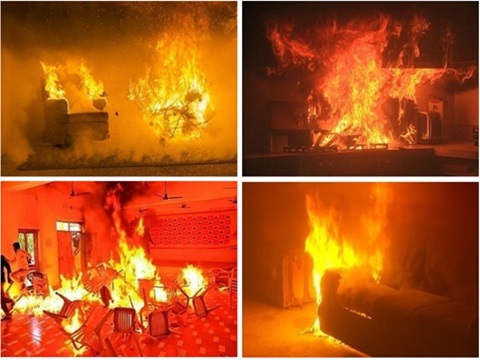}
    \end{minipage}}
    \\
    \subfloat[Air Crush]{
    \begin{minipage}{0.49\textwidth}
    \includegraphics[width=1\linewidth]{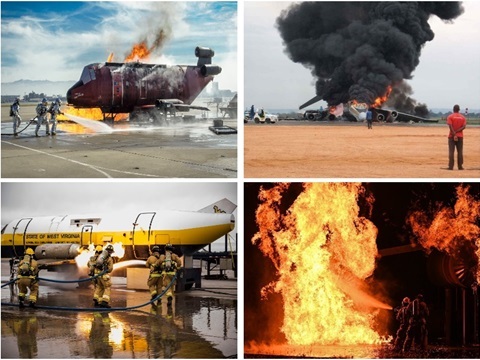}\vspace{0.5pt}
    \end{minipage}}
    \subfloat[Residential Building]{
    \begin{minipage}{0.49\textwidth}
    \includegraphics[width=1\linewidth]{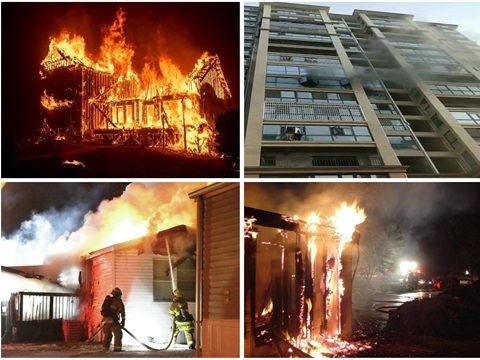}
    \end{minipage}}
\caption{Different Scenes in Our Benchmark MS-FSDB. (a) represents the Electric Power Station Scene. (b) represents the Indoor Clutter Scene. (c) represents the Air crush Scene. (d) represents the Residential Building Scene.}
\label{fig2}
\vspace{-2em}
\end{figure}

In terms of breadth, we propose a new Multi-Scene Fire and Smoke Detection Benchmark (MS-FSDB) comprising 12,586 images, depicting 2,731 scenes as illustrated in Fig.~\ref{fig1} (a). This benchmark encompasses approximately 3,603 positive examples and 8,983 negative examples. Most images within our benchmark possess dimensions exceeding 600 pixels in either length or width. Unlike previous public Fire and Smoke Detection (FSD) datasets, our benchmark not only includes flame detection but also smoke detection tasks. Additionally, it captures complex scenes featuring occlusion, multiple targets, and various viewpoints. Moreover, the dataset spans diverse environments, including residential exteriors, air crash sites, and forest settings, etc. Regarding the selection of negative examples, previous FSD datasets have often been chosen randomly and irregularly, as observed in Furg-Fire-Dataset~\cite{14}. However, our approach to selecting negative examples considers the varying difficulties posed by examples. For instance, we include images with colors resembling those of sunsets and varying lighting conditions as negative examples to standardize the detection model during the training phase. Furthermore, distinguishing between clouds and smoke is significant. In our research, clouds are classified as negative examples since they pose no threat to human life and play a crucial role in Earth's climate and water cycle~\cite{21}.
For depth, assessing the quality of FSD datasets is imperative for ensuring robust performance. In this regard, we conduct a performance analysis focusing on several salient aspects of the newly proposed MS-FSDB. Firstly, we annotate smoke and flame images with distinct labels, facilitating the training of both classification and regression tasks. Additionally, manual tagging proves more accurate than software tagging, as it enables a deeper understanding of the scene and consideration of additional details and characteristics, ultimately yielding more precise results. Secondly, while the majority of images in our benchmark are selected for their high resolution, a small portion may have undergone compression on the Internet, leading to a decrease in DPI. This scene is realistic, as we cannot guarantee that every image will maintain high DPI during transmission and use. This simulation aids in more accurately evaluating the performance of algorithms in real-world applications and better addressing various practical challenges. We choose real scene images over those generated by models because, although generative model images can mimic real-life scenes, achieving complete realism can often be challenging. Moreover, each image in our benchmark is unique, ensuring there is no redundancy, and every example showcases distinct features and information. This allows us to cover a wider and more diverse range of scenes compared to datasets containing duplicate images, enabling our benchmark to better capture diversity. This approach guarantees that even with a limited dataset, our benchmark still provides rich and varied information, encompassing a broader spectrum of real-world application scenes. For example, we show four different scenes in our benchmark MS-FSDB in Fig.~\ref{fig2}. Our benchmark maintains a positive-to-negative example ratio of approximately 1:3, aligning with traditional object detection datasets. Specifically, there are around 1,800 smoke images and 2,800 flame images, facilitating better training of the model to comprehend and handle diverse complex scenes. Finally, our benchmark surpasses the number of scenes in previous FSD datasets, comprising a total of 2,731 scenes identified through manual counting. This information is illustrated in Fig.~\ref{fig1} (a) and correlates with the total number of scenes in MS-FSDB shown in Fig.~\ref{fig1} (b). Specifically, within our benchmark, there are 96 types of fires in the commercial district scenes and 38 fires in the volcano scenes, among various other unlisted scenes. This aspect enhances the model's generalization across different scenes and enables its improved application in practical situations.\\

\section{Fire and Smoke Detection Datasets Processing}
\label{sec:fsd_datasets}
We compare our benchmark with the most popular and readily available Fire and Smoke Detection (FSD) datasets.\\

\textbf{Fire-Smoke-Dataset}~\cite{10} comprises 3,000 images, and it suffers from severe watermarking and a limited quantity of images. We randomly segment it to create a 2,100 training set and a 900 testing set, and label this dataset for location.\\

\textbf{Furg-Fire-Dataset}~\cite{14} consists of 23 videos. These videos are divided into individual frames to form images. The computer randomly deletes redundant photos while retaining 3,000 images and labeling them accordingly. However, it should be noted that these images are not clear due to the presence of simulated scenes and the absence of real scenes.\\

\textbf{VisiFire}~\cite{11} is a public FSD dataset consisting of 40 videos. The processing principle remains consistent with our earlier method mentioned, while resulting in 2,500 retained images. It is important to note that VisiFire exhibits low image resolution and a limited variety of scene compositions. We choose 1,750 images for the training set and 750 images for the testing set, and label them accordingly.\\

The dataset utilized in our research is \textbf{FIRESENSE}~\cite{12}, which comprises 49 videos. The processing methodology adopted in this study aligns with that of VisiFire, involving the retention of 2,500 images for analysis. Regression labels are assigned to those images. It is crucial to acknowledge that FIRESENSE~\cite{12} exhibits low resolution and watermarking, potentially influencing the learned features during model training.\\

\textbf{BoWFire}~\cite{13} consists of 240 images in the training set, as well as 226 images in the dataset. However, the training set images have low resolution, making them unsuitable for comparison with current detection models. Consequently, they are removed, and only the dataset containing 226 images is retained. Nevertheless, the dataset quantity is relatively small compared to modern standards. We choose 158 images for the training set and 68 images for the testing set, labeling them accordingly.\\

For \textbf{our proposed benchmark}, to ensure experimental fairness, we randomly select 3,000 images from our proposed benchmark (referred to as miniMS-FSDB). This choice is made because our benchmark significantly exceeds the size of previous public FSD datasets. Our benchmark (referred to as MS-FSDB) consists of 12,586 images, with 8,983 images for training, 3,528 images for testing, and 11,327 images for the trainval subset. Furthermore, the benchmark is compared to the public FSD dataset, while also presenting the results obtained using the complete benchmark.\\

To objectively evaluate the generalization performance of the FSD dataset, we create a new \textbf{testing set} consisting of one hundred randomly selected labeled images from all FSD testing sets we used.\\

\section{Baselines}
\label{sec:baselines}
We compare our proposed Fire and Smoke Detection (FSD) method with several representative and general object detection baselines.\\

Single Shot MultiBox Detector (\textbf{SSD})~\cite{22} algorithm is a renowned object detection method in the industry,  renowned for its one-stage detection without the need for complex region proposals. Through one forward pass, it swiftly generates object category probabilities and precise locations, striking a balance between speed and accuracy.   Furthermore, SSD's multi-scale detection enhances its ability to identify objects of diverse sizes, expanding the detection field of view and optimizing performance. The input image sizes for SSD(s/l) model are $300~\times~300$ for small (s) images and $512~\times~ 512$ for large (l) images. \\

\textbf{RetinaNet}~\cite{23} is an advanced object detection algorithm that combines Backbone Networks (e.g., ResNet~\cite{24}) with Neck Networks (e.g.,Feature Pyramid Networks (FPN)~\cite{25}) to capture multi-scale contextual information for addressing various object detection tasks. Notably, RetinaNet introduces the innovative Focal Loss function to effectively mitigate common class imbalance issues in object detection, thereby improving detection accuracy for small and occluded objects. For RetinaNet(s/l) model, the input image sizes are approximately $360~\times~360$ for “s” and approximately $608~\times~608$ for “l”.\\

\textbf{Faster RCNN}~\cite{26} algorithm is an efficient method for object detection that achieves precise results by combining feature extraction, region proposal network, ROI pooling, and classification-regression head network. The core of this algorithm is the Region Proposal Network (RPN), which rapidly generates candidate regions for detailed classification and localization, making a significant impact in the field of object detection. In the case of Faster RCNN (s/l) model, the input image sizes are around $400~\times~666$ for “s” and around $800~\times~1,333$ for “l”. \\

Fully Convolutional One-Stage Object Detection (\textbf{FCOS})~\cite{27} is an innovative algorithm for object detection that predicts object categories and locations pixel by pixel using a fully convolutional network and incorporates object centrality for improved localization accuracy. It distinguishes itself from traditional object detection methods by achieving anchor-free detection. As for FCOS(s/l) model, the input image sizes are roughly $416~\times~608$ for “s” and about $832~\times~1,216$ for “l”.\\

\textbf{YOLOv5}~\cite{28} is an advanced and efficient object detection algorithm that showcases outstanding performance and innovative model design. It utilizes lightweight networks (e.g., CSPDarkNet53~\cite{29}) for feature extraction to ensure efficient detection. The core strength lies in its multi-scale detection mechanism, utilizing feature pyramids~\cite{25} to handle objects of various sizes, thereby enhancing detection accuracy. For handling YOLOv5 image sizes, we only use pixel values of $640~\times~640$. \\

Previous object detection algorithms, while versatile, have shown poor performance in FSD tasks. To address this issue, we propose a novel algorithm optimized for FSD tasks (\textbf{our method} referred to Ours). By retaining the feature extraction and integration advantages of general detection algorithms, we specifically redesign the fire-related target detection head network to cater to the unique requirements of FSD. As this paper focuses on introducing benchmark, detailed model specifics will not be extensively discussed. In our method(s/l) model, the input image sizes are roughly 416 × 608 for “s” and about 832 × 1, 216 for “l”.\\

\section{Experiments and Analysis}
\label{sec:experiments}
\subsection{Setting}
We successfully reproduce SSD~\cite{22}, RetinaNet~\cite{23}, Faster RCNN~\cite{26}, FCOS~\cite{27}, YOLOv5~\cite{28}, and our proposed model using the PyTorch framework. Experiments are conducted on multiple public FSD datasets, including
VisiFire~\cite{11}, Fire-SmokeDataset~\cite{10}, Furg-Fire-Dataset~\cite{14}, FIRESENSE~\cite{12}, BoWFire~\cite{13}, and our benchmark MS-FSDB. To ensure consistency and comparability, all model codes are standardized, and all datasets are uniformly preprocessed and converted to standardized formats. The experiments are performed on a computing platform equipped with an NVIDIA RTX 3090Ti GPU to ensure efficient computational capabilities and accurate experimental results.\\

We establish uniform evaluation criteria and utilize the 11-point interpolation method~\cite{30} to accurately measure the performance of the model. For each dataset in the section “Basic Experiments”, we calculate the average precision of flame and smoke categories separately, and then compute the overall average precision across all categories to comprehensively and objectively assess the model’s detection performance.

\subsection{Basic Experiments}
According to Table~\ref{tab2}, the baselines SSD, RetinaNet, Faster RCNN, and FCOS achieve mean Average Precision (mAP) scores of 89.4, 89.6, 96.6, and 96.3, respectively, on MS-FSDB with large-sized images. This indicates that the detection accuracy of our anchor-based one-stage object detection algorithm is lower than that of the anchor-based two-stage object detection algorithm and anchor-free detection algorithm, when trained and tested on the same datasets. This difference could be attributed to the generation of a large number of anchors during the computation process of the anchor-based object detection algorithm, inevitably introducing some background noise. However, the anchor-based two-stage object detection algorithm adopts a different strategy to handle background noise, thereby improving detection accuracy. Furthermore, the baselines SSD (mAP 77.8) and RetinaNet (mAP 84.9) models on miniMS-FSDB with small-sized images do not exhibit higher mAP than the SSD model (mAP 84.1) and RetinaNet model (mAP 90.7) on the Fire-Smoke-Dataset. This indirectly proves the challenging nature of our benchmark. When our miniMS-FSDB is small and our model is complex, our model may struggle to extract target features, resulting in poor performance on our miniMS-FSDB and potentially leading to overfitting. Conversely, when our MS-FSDB is large and our model is complex, our model can access more diverse and representative samples, thereby enhancing its generalization ability to our benchmark. It is worth noting that the BoWFire dataset is unique with 226 images, and its low accuracy is due to overfitting of the dataset.\\

\begin{figure}[!htb]
    \centering
    \subfloat[False results of general detection]{
    \includegraphics[width=0.24\linewidth]{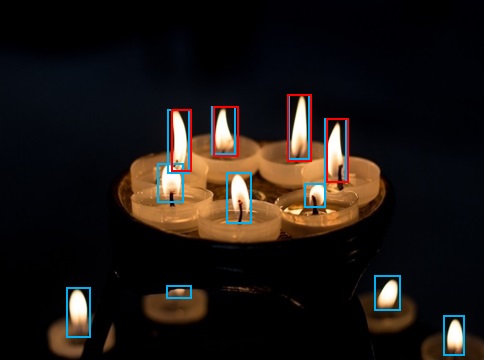}\vspace{1pt}
    \includegraphics[width=0.24\linewidth]{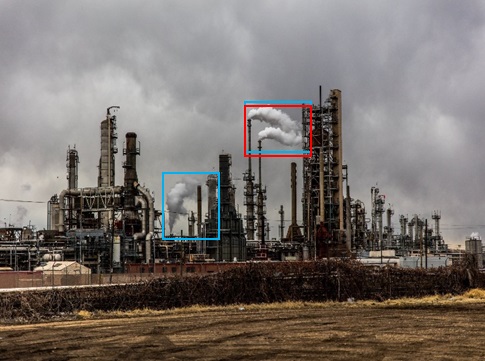}\vspace{1pt}
    \includegraphics[width=0.24\linewidth]{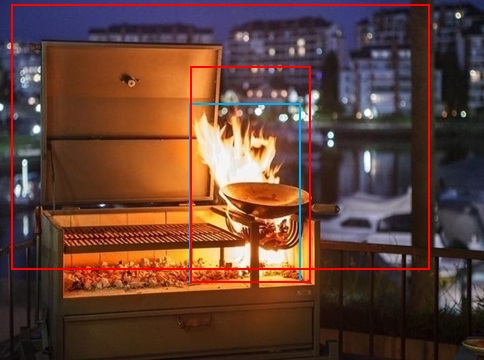}\vspace{1pt}
    \includegraphics[width=0.24\linewidth]{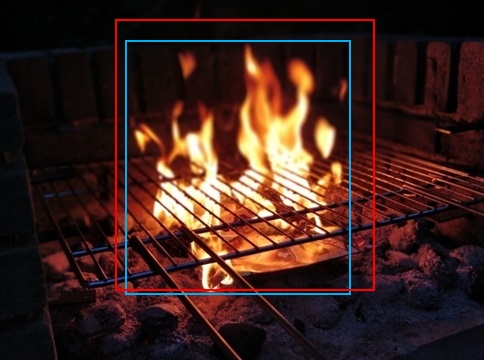}\vspace{1pt}}
    \\
    \subfloat[Correct results of our method]{
    \includegraphics[width=0.24\linewidth]{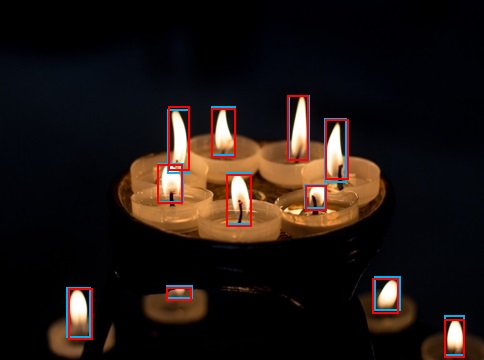}\vspace{1pt}
    \includegraphics[width=0.24\linewidth]{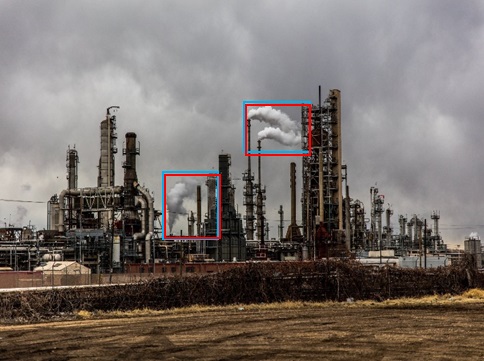}\vspace{1pt}
    \includegraphics[width=0.24\linewidth]{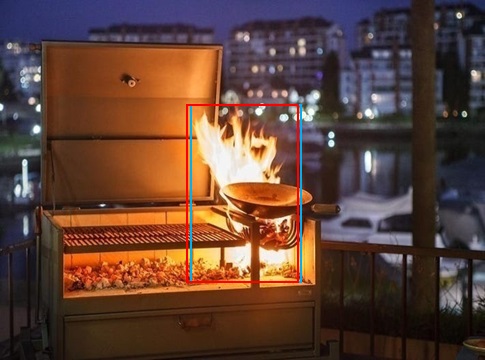}\vspace{1pt}
    \includegraphics[width=0.24\linewidth]{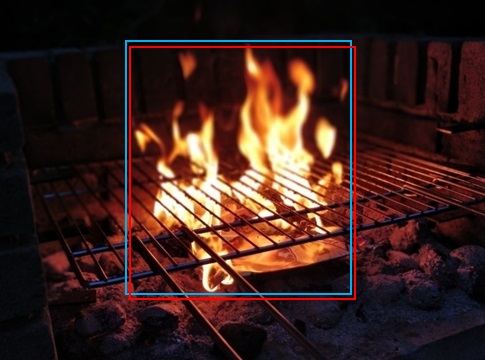}\vspace{1pt}}
\caption{For ordinary fire and smoke, some of the difficulties encountered in detection are addressed using our method. (a) represents the false results of the general detection. (b) represents the correct results of our method. In the diagram, blue boxes represent ground truth, and red boxes represent predicted results.}
\label{fig3}
\end{figure}

As shown in Fig.~\ref{fig3} (a), we can observe that the general detectors face various challenges. From left to right, these include: the first image exhibiting the issue of missing detection of multiple small objects; the second image demonstrating difficulty in distinguishing RGB similarity between the foreground (smoke) and background; the third image showing failure to identify the RGB similarity between the foreground (fire) and background; and the fourth image revealing low accuracy of regression.\\

\begin{sidewaystable}
        \renewcommand\arraystretch{0.3}
        \centering
        \begin{minipage}{\textwidth}
        \caption{Baseline models comparison across different datasets, where Fire represents the Average Precision ($\%$) of fire, Smoke represents the Average Precision ($\%$) of smoke, and $\mathrm{mAP}$ represents the mean Average Precision ($\%$) of fire and smoke. “s" represents the input image of small size, while “l" represents the input image of large size, and the specific content is in the subsection “Basic Experiments". F-RCNN means Faster RCNN. Ours means our method.}
        \label{tab2}
        \begin{tabularx}{\textwidth}{XXXXX | XXX | XXX | XXX |XXX}
            \toprule
            \diagbox{Dataset}{AP ($\%$)}{Model}& & \multicolumn{3}{c}{SSD\textsuperscript{\cite{22}}(s/l)} & \multicolumn{3}{c}{RetinaNet\textsuperscript{\cite{23}}(s/l)}&
            \multicolumn{3}{c}{F-RCNN\textsuperscript{\cite{26}}(s/l)}& \multicolumn{3}{c}{FCOS\textsuperscript{\cite{27}}(s/l)}& \multicolumn{3}{c}{Ours(s/l)} \\
            
            \midrule
            && \rmfamily{Fire}& \rmfamily{Smoke}& $\mathrm{mAP}$& \rmfamily{Fire}& \rmfamily{Smoke}& $\mathrm{mAP}$& \rmfamily{Fire}& \rmfamily{Smoke}& $\mathrm{mAP}$& \rmfamily{Fire}& \rmfamily{Smoke}& $\mathrm{mAP}$&
            \rmfamily{Fire}& \rmfamily{Smoke}& $\mathrm{mAP}$\\
            \multicolumn{2}{l}{Fire-Smoke-Dataset\textsuperscript{\cite{10}}(s)}&77.5&90.8&84.1&
            90.7&90.7&90.7&97.3&93.2&95.3&97.2&98.5&97.8&\textbf{97.5}&\textbf{99.2}&\textbf{98.3}\\
            \multicolumn{2}{l}{Furg-Fire-Dataset\textsuperscript{\cite{14}}(s)}&75.8&86.9&81.4&
            81.7&90.0&85.8&\textbf{95.8}&93.4&94.6&93.7&\textbf{98.1}&95.9&94.2&\textbf{98.1}&\textbf{96.1}\\
            \multicolumn{2}{l}{VisiFire\textsuperscript{\cite{11}}(s)}&78.2&89.5&83.9&
            84.2&90.7&87.4&92.8&88.8&90.8&88.8&96.7&92.8&\textbf{96.2}&\textbf{99.4}&\textbf{97.8}\\
            \multicolumn{2}{l}{FIRESENSE\textsuperscript{\cite{12}}(s)}&89.0&90.4&89.7&
            90.9&90.9&90.9&96.8&95.8&96.3&96.1&96.1&96.1&\textbf{98.3}&\textbf{98.1}&\textbf{98.2}\\
            \multicolumn{2}{l}{BoWFireDataset\textsuperscript{\cite{13}}(s)}&69.6&84.7&77.1&
            72.3&88.4&80.3&\textbf{86.3}&95.0&90.6&85.1&95.2&90.2&86.1&\textbf{97.9}&\textbf{92.0}\\
            \midrule
            \multicolumn{2}{l}{miniMS-FSDB(s)}&71.2&84.4&77.8&
            80.4&89.4&84.9&98.0&93.0&95.5&94.1&95.9&95.0&\textbf{98.3}&\textbf{99.3}&\textbf{98.8}\\
            \multicolumn{2}{l}{MF-FSDB(s)}&81.0&90.2&85.6&
            81.0&90.5&85.8&\textbf{98.2}&93.5&95.8&95.6&98.4&97.0&97.1&\textbf{98.9}&\textbf{98.0}\\
            \multicolumn{2}{l}{miniMS-FSDB(l)}&75.9&87.1&81.5&
            87.0&88.4&87.7&98.0&94.1&96.1&95.5&96.2&95.8&\textbf{98.1}&\textbf{97.6}&\textbf{97.9}\\
            \multicolumn{2}{l}{MS-FSDB(l)}&88.0&90.9&89.4&
            89.5&89.6&89.6&97.3&95.9&96.6&96.0&96.7&96.3&\textbf{98.4}&\textbf{98.6}&\textbf{98.5}\\
            \bottomrule
        \end{tabularx}
        \end{minipage}

        \hfill
        \renewcommand\arraystretch{0.3}
        \centering
        \begin{minipage}{\textwidth}
        \caption{Cross experimental results, where results represent the mean Average Precision ($\%$) of fire and smoke. the table on the right, we train baseline models using the proposed miniMS-FSDB and then test with previous FSD dataset. For example, a mAP of 45.8 represents the result of the SSD model trains on our miniMS-FSDB and then test using the Fire-Smoke-Dataset. The table on the left, we train baseline models using previous FSD dataset and then test with miniMS-FSDB. For example, a mAP of 45.5 represents the result of the SSD model trains on Fire-Smoke-Dataset and the test using our miniMS-FSDB. We uniformly choose small input images for all baseline models to ensure consistency and comparability in experiments. F-RCNN means Faster RCNN and Ours means our method.}  
        \label{tab3}
        \begin{tabularx}{\textwidth}{XX XXXXX XXXXX}  
        \toprule  
        \multicolumn{2}{l}{\diagbox{mAP ($\%$)}{Model}{TestDataset}} & \multicolumn{4}{c}{\color{red} miniMS-FSDB} & \multicolumn{4}{c}{\color{red}Other FSD Datasets} \\  
        \cmidrule(lr){3-7} \cmidrule(lr){8-12}  
        &Dataset&\footnotesize{SSD}&\footnotesize{RetinaNet} &\footnotesize{F-RCNN}&\footnotesize{FCOS}&\footnotesize{Ours}&\footnotesize{SSD}&\footnotesize{RetinaNet}&\footnotesize{F-RCNN}&\footnotesize{FCOS}&\footnotesize{Ours} \\  
        \midrule  
        \multicolumn{2}{l}{\color{blue}Fire-Smoke-Dataset\textsuperscript{\cite{10}}}&\textbf{45.8}&\textbf{16.6}&\textbf{46.3}&\textbf{40.9}&\textbf{41.6}&45.5&7.0&6.2&36.7&36.3 \\  
        \multicolumn{2}{l}{\color{blue}Furg-Fire-Dataset\textsuperscript{\cite{14}}}&\textbf{42.1}&\textbf{20.2}&\textbf{40.8}&\textbf{37.9}&\textbf{39.8}&28.6&2.0&17.1&8.2&18.0\\  
        \multicolumn{2}{l}{\color{blue}VisiFire\textsuperscript{\cite{11}}}&17.6&\textbf{10.3}&\textbf{14.6}&11.8&11.4&\textbf{33.8}&1.0&6.0&\textbf{16.9}&\textbf{18.2}\\  
        \multicolumn{2}{l}{\color{blue}FIRESENSE\textsuperscript{\cite{12}}}&\textbf{55.6}&\textbf{8.3}&\textbf{60.1}&\textbf{54.1}&5.2&29.9&4.7&13.9&15.1&\textbf{21.2}\\  
        \multicolumn{2}{l}{\color{blue}BoWFireDataset\textsuperscript{\cite{13}}}&12.4&\textbf{10.4}&\textbf{14.5}&\textbf{15.8}&\textbf{14.4}&\textbf{19.3}&1.4&6.2&3.4&5.5\\   
    \bottomrule  
    \end{tabularx}
    \end{minipage}
\end{sidewaystable}

\subsection{Performance Verification}
We employ two cross-validation methods in our study. We uniformly choose small input images for all baseline models to ensure consistency and comparability in experiments. First, we train baseline models on our miniMS-FSDB and then test them on the previous FSD dataset in Table~\ref{tab3} on the right. For example, a mAP of 45.8 represents the result of the SSD model trains on our miniMS-FSDB and then test using the Fire-Smoke-Dataset. Second, we train the baseline models on the previous FSD dataset and then test them on our miniMS-FSDB in Table~\ref{tab3} on the left. For example, a mAP of 45.5 represents the result of the SSD model trains on the Fire-Smoke-Dataset and then test using our miniMS-FSDB. This cross-validation experimental design yields satisfactory results, confirming the excellent generalization capabilities of our benchmark dataset.\\

Although RetinaNet performed well on various object detection tasks, its performance was limited on our fire and smoke detection dataset, especially in cross-training environments. The challenges posed by complex textures and reflective properties, such as the dynamic nature of fire and the semi-transparency of smoke, significantly impacted  performance of RetinaNet. In contrast, SSD, Faster R-CNN, YOLOv5, and FCOS models exhibited superior adaptability and stability in cross-training settings, effectively handling the complexities of the datasets due to their unique strengths and capabilities.\\

We carefully construct a new testing set comprising one hundred labeled images randomly selected from all FSD testing sets we used, to ensure a comprehensive and unbiased evaluation. The experimental results demonstrate that our benchmark exhibits stronger generalization abilities in FSD tasks compared to previous FSD datasets in Table~\ref{tab4}. Specifically, our model trained on the MS-FSDB dataset achieved an optimal performance accuracy of 75.4, on the new testing set. 

\begin{table}
    \vspace{-1em}
\renewcommand\arraystretch{0.5}
        \caption{We train baselines on different FSD datasets and then test with the new testing set, where $\mathrm{mAP}$ represents the mean Average Precision ($\%$) of fire and smoke. We use the input image of small size. F-RCNN means Faster RCNN, and Ours means our method.}
        \label{tab4}
        \begin{tabularx}{\textwidth}{XXc | c | c | c |c}
            \toprule
            \multicolumn{2}{l}{\diagbox{mAP ($\%$)}{Model}{TestDataset}}& 
            \multicolumn{5}{c}{\color{red}Testing Set}\\
             \cmidrule(lr){3-7}&Dataset&
             \multicolumn{1}{c}{SSD\textsuperscript{\cite{22}}}&
             \multicolumn{1}{c}{RetinaNet\textsuperscript{\cite{23}}}&
             \multicolumn{1}{c}{F-RCNN\textsuperscript{\cite{26}}}& 
             \multicolumn{1}{c}{FCOS\textsuperscript{\cite{27}}}&
             \multicolumn{1}{c}{Ours} \\
            
            \midrule
            \multicolumn{2}{l}{\color{blue}Fire-Smoke-Dataset\textsuperscript{\cite{10}}}&\textbf{42.7}&8.9&40.9&37.2&36.8\\
            \multicolumn{2}{l}{\color{blue}Furg-Fire-Dataset\textsuperscript{\cite{14}}}&\textbf{27.5}&2.3&16.2&10.0&18.4\\
            \multicolumn{2}{l}{\color{blue}VisiFire\textsuperscript{\cite{11}}}&\textbf{31.1}&1.2&9.4&20.6&21.8\\
            \multicolumn{2}{l}{\color{blue}FIRESENSE\textsuperscript{\cite{12}}}&\textbf{26.9}&10.7&13.1&15.0&22.0 \\
            \multicolumn{2}{l}{\color{blue}BoWFireDataset\textsuperscript{\cite{13}}}&\textbf{28.3}&5.5&16.0&16.3&19.3\\
            \multicolumn{2}{l}{\color{blue}miniMS-FSDB}&69.4&12.6&68.0&66.6&\textbf{74.7}\\
            \multicolumn{2}{l}{\color{blue}MS-FSDB}&66.4&8.2&70.0&74.5&\textbf{75.4}\\
            \bottomrule
        \end{tabularx}
         \vspace{-1.5em}
\end{table}

\begin{table}[!htb]
    \caption{Performance comparison of different models on our miniMS-FSDB. Key metrics include detection accuracy, mAP, Performance (Gflops), and number of parameters (M) for each category. Fire represents the Average Precision (\%), Smoke represents the Average Precision (\%), and mAP represents the mean Average Precision (\%) of fire and smoke. For consistency and comparability, we use small input images for all baseline models. Ours denotes our method.}
    \label{tab5}
    \centering
    \renewcommand\arraystretch{0.8}
    \begin{tabularx}{\textwidth}{l *{5}{>{\centering\arraybackslash}X}}
        \toprule
        Model & Fire (\%) & Smoke (\%) & mAP (\%) & Perf (Gflops) & Params (M) \\
        \midrule
        YOLOv5s\textsuperscript{\cite{28}} & 66.3 & 81.3 & 73.8 & 15.8 & 7.0 \\
        YOLOv5x\textsuperscript{\cite{28}} & 67.5 & 81.6 & 74.5 & 203.8 & 86.2 \\
        YOLOv5n\textsuperscript{\cite{28}} & 72.4 & 78.2 & 75.3 & 4.1 & \textbf{1.8} \\
        YOLOv5l\textsuperscript{\cite{28}} & 71.5 & 79.7 & 75.6 & 107.7 & 46.1 \\
        YOLOv5m\textsuperscript{\cite{28}} & 69.6 & 84.1 & 76.8 & 47.9 & 20.9 \\
        SSD\textsuperscript{\cite{22}} & 71.2 & 84.4 & 77.8 & 731.5 & 23.9 \\
        RetinaNet\textsuperscript{\cite{23}} & 80.4 & 89.4 & 84.9 & \textbf{1,715.3} & 36.4 \\
        FasterRCNN\textsuperscript{\cite{26}} & 98.0 & 93.0 & 95.5 & 504.3 & 41.4 \\
        FCOS\textsuperscript{\cite{27}} & 94.1 & 95.9 & 95.0 & 1,289.9 & 32.1 \\
        Ours & \textbf{98.3} & \textbf{99.3} & \textbf{98.8} & 1,290.5 & 33.3 \\
        \bottomrule
    \end{tabularx}
\end{table}

\subsection{Summary}
We compare various FSD models on our miniMS-FSDB. And we uniformly choose small input images for all baseline models to ensure consistency and comparability in experiments. By analyzing Table~\ref{tab5} in depth, we can observe the following important findings: Firstly, the benchmark we designed is carefully crafted to address the complexity and diversity of the problem, aiming to provide a fair and challenging training environment for multiple baseline methods, including ours. Training on this benchmark enables baseline methods to undergo more accurate and comprehensive performance testing on a new testing set. Specifically, our method's AP for Fire reaches 98.3, and for Smoke, the AP also achieves a strong performance of 99.3. Overall, our method's mAP is high at 98.8, Compared to various other FSD datasets, our benchmark demonstrates superior generalization performance in Table~\ref{tab4}, offering researchers a more reliable and comprehensive platform for performance evaluation purposes. Furthermore, our model successfully corrects the problems present in the previous generic detection model, as illustrated in Fig.~\ref{fig3}. From left to right, the first image represents the missing detection of multiple small objects. The second image represents the failure to identify the RGB similarity between foreground (smoke) and background, and the third image represents the failure to identify the RGB similarity between foreground (fire) and background. Additionally, the fourth image demonstrates low regression errors.

\section{Conclusion}
We propose a new Multi-Scene Fire and Smoke Detection Benchmark (MS-FSDB) that expands a large number of real fire or smoke scenes environments and complex fire scenes compared with previous FSD datasets. To assess the generalization of our benchmark, we employ a comprehensive experimental approach. Initially, we conduct cross-validation experiments to ensure an unbiased evaluation. Furthermore, to enhance the fairness of the assessment, we specifically create a new testing set dedicated to experimenting with all FSD testing sets. The experimental findings unequivocally demonstrate that, compared to other datasets, our benchmark exhibits superior generalization performance in the FSD task.

\section*{Acknowledgements}
This work is supported by the Fundamental Research Funds for the Central Universities (226-2024-00058), the National Natural Science Foundation of China (Grant No. 62106235), and Zhejiang Province "Pioneering Soldier" and "Leading Goose" R\&D Project (2023C01027). 

\bibliographystyle{unsrt}  

\bibliography{references}
\end{document}